\documentclass[]{spie}  %>>> use for US letter paper
\usepackage{comment}
\usepackage{booktabs}
\usepackage{seqsplit}
\usepackage{url}
\usepackage{amsmath,amsfonts,amssymb}
\usepackage{graphicx}
\usepackage{gensymb}
\usepackage{rotating}  % for \rotatebox
\usepackage{caption}
\usepackage{subcaption}
\usepackage{enumitem}
\usepackage{xcolor}
\usepackage{float}
\usepackage[colorlinks=true, allcolors=blue]{hyperref}
\usepackage{listings}

\title{Domain generalization and synthetic data in object detection: the enabler, the probe, and the gap}

\author[a]{Elfi I.S. Hofmeijer}
\author[a]{Ella P. Fokkinga}
\author[a]{Friso G. Heslinga}
\author[a]{Klamer Schutte}
\author[b]{Jörgen M. Karlholm}
\affil[a]{TNO - Defence, Security and Safety, the Hague, the Netherlands}
\affil[b]{FOI - Swedish Defence Research Agency, Linköping, Sweden}
\authorinfo{Corresponding author: Friso G. Heslinga. E-mail: fgheslinga@gmail.com}

\begin{document} 
\maketitle

\begin{abstract}
% In niche domains such as the military, acquiring large and representative real-world image datasets for developing object detectors is often infeasible, difficult, or costly. The use of synthetic data offers a promising alternative to increase the number and variation of images. Still, object detection models often suffer from performance degradation when deployed on real imagery due to domain shifts. Domain Generalization (DG) aims to train models so that they remain robust under distribution shifts and generalize well to unseen domains, making synthetic-to-real transfer a central DG problem. DG research focused on object detection is scarce, although these models face specific challenges such as global scene understanding and localization across different spatial scales. This paper presents an object detection-centric review of DG methods, with a two-fold perspective on the role of synthetic data. First, we review how synthetic data can be used to enable DG, where existing approaches are organized along two cornerstones: (1) diversification, which focusses on enlarging support of the training distribution, and (2) alignment, which centers around constraining the predictions or representations of the model to be more domain-invariant. Second, we discuss how synthetic data can be leveraged to analyze failures of DG, by deliberately creating images in different domains or perturbations. This review highlights existing results of both approaches and suggests research directions towards more object detection-focused DG methods. 

Object detection models often experience performance degradation when deployed under distribution shifts, caused by for example changes in weather type, operational environment, or object appearance. Domain Generalization (DG) aims to develop models that remain robust under such shifts and generalize well to unseen domains. DG research specifically focused on object detection models is scarce, although these models face additional challenges around localization and multi-scale representations. Synthetic data is a promising tool to support in DG, by enabling large-scale generation of diverse new samples. In this paper, we present an object detection-centric review of DG and examine the role of synthetic data from three complementary perspectives. First, synthetic data acts as an \textit{enabler} of DG through diversification and alignment strategies that aim to improve robustness to distribution shifts. Second, it serves as a \textit{probe} that enables controlled experimentation to identify and understand failure modes. Third, we discuss the synthetic-to-real \textit{gap}, a particularly challenging form of domain shift that arises when models trained on synthetic imagery are deployed on real-world data. Through reviewing these perspectives, we identify limitations of current DG approaches for object detection and argue that future research requires representation-aware methods that explicitly address both localization and classification under domain shift.

\end{abstract}

\keywords{Domain generalization; Synthetic data; Object detection}

\section{Introduction}
\label{sec:intro} 

Object detection systems are often deployed in environments that differ substantially from the training domain, especially in domains where limited data is available. Changes in weather conditions \cite{zhang2026rtsdgodrealtimesingledomaingeneralized}, image quality \cite{becker2026selfawareobjectdetectiondegradation}, or operational environment \cite{9093355} can all lead to degraded detection performance due to a shift in distribution. Even large foundation models exhibit performance drops when confronted with out-of-domain (OOD) data, especially in niche domains, highlighting the importance of tackling the challenge of robust generalization. \cite{Mao2023}

% Even though state-of-the-art object detection models can achieve high performance, the performance drops often in real-world scenarios. High object detection performance is achieved by pre-training large models on large, diverse datasets, followed by careful fine-tuning on the downstream task. However, it should be noted that there is still a very significant performance drop for Out-of-Domain (OOD) data, even for the largest models. Mao et al. \cite{Mao2023} have investigated the robustness to domain shifts for many object detection algorithms. All models experience a large drop in performance for OOD data, but high performance on in-domain data typically results in relatively high performance on OOD data. The best performance is achieved by recent giant foundation models trained on billions of images. However, if these models are trained on smaller datasets (e.g., COCO), their robustness decreases dramatically.

% (Yu, et al., 2021) study empirically how pre-training affects OOD generalization of convolutional networks (convnets) for image classification, and find that larger and more diverse pre-training datasets result in better in-domain and OOD performance, as does larger model size.

% SOLUTION -> domain generalization 
The goal of Domain Generalization (DG) is to train models that remain robust under distribution shifts and generalize well to previously unseen domains. Despite DG being an active research field, DG methods for object detection are still scarce. Comprehensive surveys exist for DG in general computer vision \cite{Zhou2023,Wang2023}, but only a small fraction of the literature addresses the unique challenges posed by object detection. Unlike image classification, object detection requires both global scene understanding for localization and instance-level reasoning for classification, making many DG methods developed for classification difficult to transfer directly. These challenges are particularly pronounced in niche domains, such as the military domain, where acquiring large and representative datasets is often expensive or infeasible. Limited training data increases the likelihood of distribution shifts between training and deployment conditions, often resulting in substantial performance degradation when object detectors are applied in real-world scenarios.

% Two key concepts in DG are diversification and alignment. Training the model on a diverse dataset may prevent it from focusing on domain-specific “spurious” features (e.g., texture) and instead make it base its decisions on more robust features (e.g., shape). Explicitly aligning outputs and features for data from different domains is a more direct way of enforcing domain invariance. When models are trained on largely synthetic data, the DG problem also becomes a synthetic-to-real transfer problem. In that case it is imperative that a model trained on synthetic images, generalizes to the real-world variants.
% DG for object detection is lacking --> daarom deze review 

% synthetic data
% A promising alternative to increase the amount of images is by creating synthetic or generated versions from varying scenarios. The use of synthetic data bypasses the difficulty and costs of collecting and annotating sufficient amounts of real imagery with objects of interest,

Synthetic data offers a promising tool to increase both the quantity and diversity of data, introducing variation in factors such as environments, weather conditions, object appearance, and sensor characteristics. Synthetic imagery can be created through a range of methodologies that differ in their underlying data generation process, including graphics-based simulation \cite{gaidon2016virtual, Eker2023}, image compositing \cite{ghiasi2021simple, guo2024enrich}, and, more recently, foundation-model-driven generative approaches. Generative models can synthesize realistic imagery directly from textual, visual, or structural conditioning signals, enabling scalable generation of diverse training data without requiring detailed 3D assets or scene engineering \cite{fokkinga2025generative, Fokkinga2026-or}. As a result, synthetic data is increasingly used to supplement scarce datasets and expose models to a broader range of operational conditions than can typically be captured through real-world data alone.

\begin{figure}
    \centering
    \includegraphics[width=0.95\linewidth]{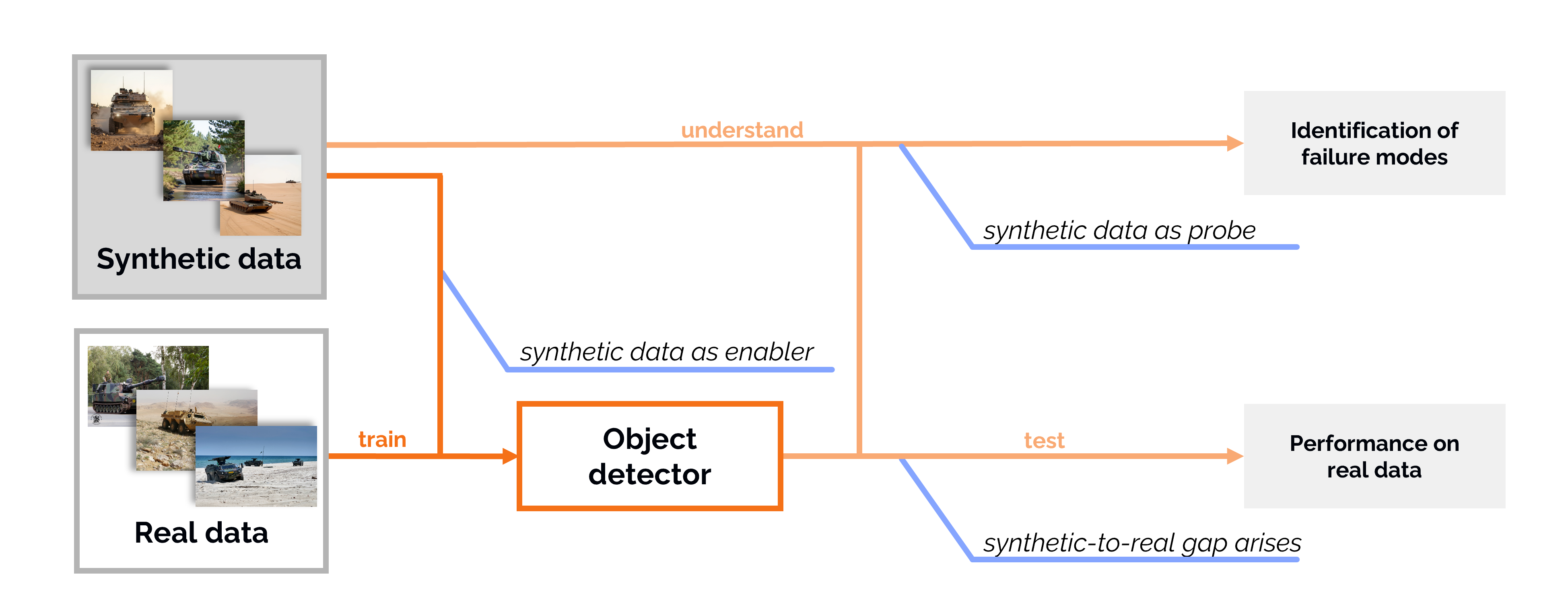}
    \caption{Overview of the three views on synthetic data in domain generalization. Synthetic data can act as an \textit{enabler} by  augmenting training data to improve domain generalization, and as a \textit{probe} by enabling controlled experiments for identifying and understanding failure modes. At the same time, the use of synthetic data may introduce a synthetic-to-real \textit{gap}, where differences between synthetic training data and real deployment data lead to performance degradation.}
    \label{fig:overview_dg}
\end{figure}

% OUTLINE OF THE PAPER 
In this paper, we first review DG from an object detection perspective, focusing on two central principles for many DG approaches: diversification and alignment. Diversification aims to expand the support of the training distribution, whereas the goal of alignment is to learn representations that remain invariant across domains. We then examine the role of synthetic data from three complementary views (Figure \ref{fig:overview_dg}). First, synthetic data can act as an \textit{enabler} of DG by supporting diversification and alignment strategies that improve robustness under distribution shifts. Second, synthetic data can serve as a \textit{probe}, enabling controlled experiments that help identify failure modes and characterize model behavior under domain shifts. Third, an important consequence of using synthetic data is the emergence of a synthetic-to-real \textit{gap}, that arises when models trained on synthetic imagery are deployed on real-world data. As illustrated in Figure \ref{fig:gap}, differences between the training and deployment distributions can lead to performance degradation, making synthetic-to-real transfer a practically relevant instance of domain shift and a useful case study for DG.

Through these three perspectives, we discuss how synthetic data can be used to improve DG, how it can be used to understand failures of generalization, and how it gives rise to synthetic-to-real transfer challenges. By reviewing existing approaches and observations from the literature, we identify key limitations of current DG methods for object detection and highlight opportunities for future research.

\section{Domain generalization}
\label{sec:domaingeneralization}
\begin{figure}
\centering 
\begin{subfigure}[t]{0.48\textwidth} 
\centering 
\includegraphics[width=\linewidth]{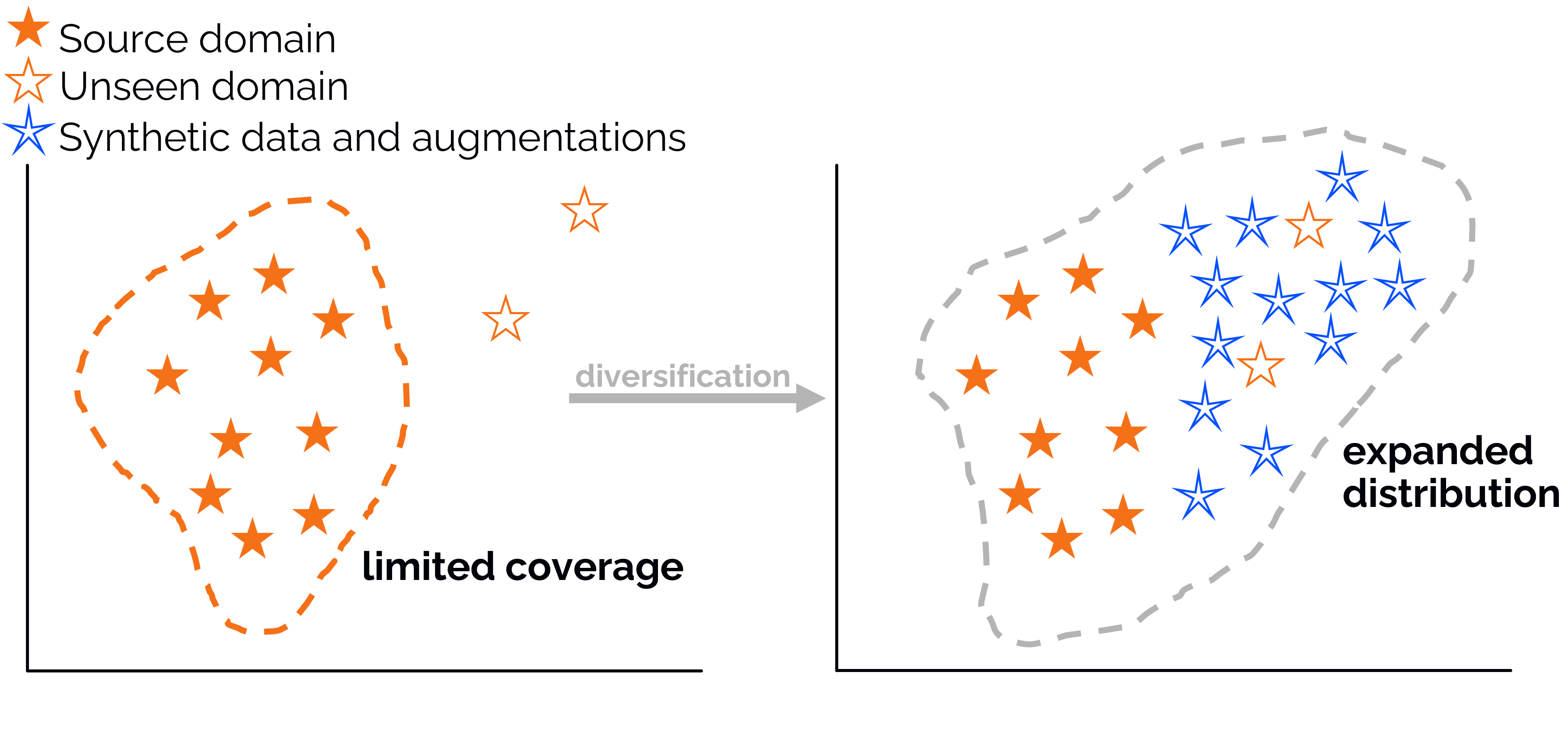} 
\caption{Diversification.} 
\label{fig:diversification} 
\end{subfigure} 
\hfill 
\begin{subfigure}[t]{0.48\textwidth} 
\centering
\includegraphics[width=\linewidth]{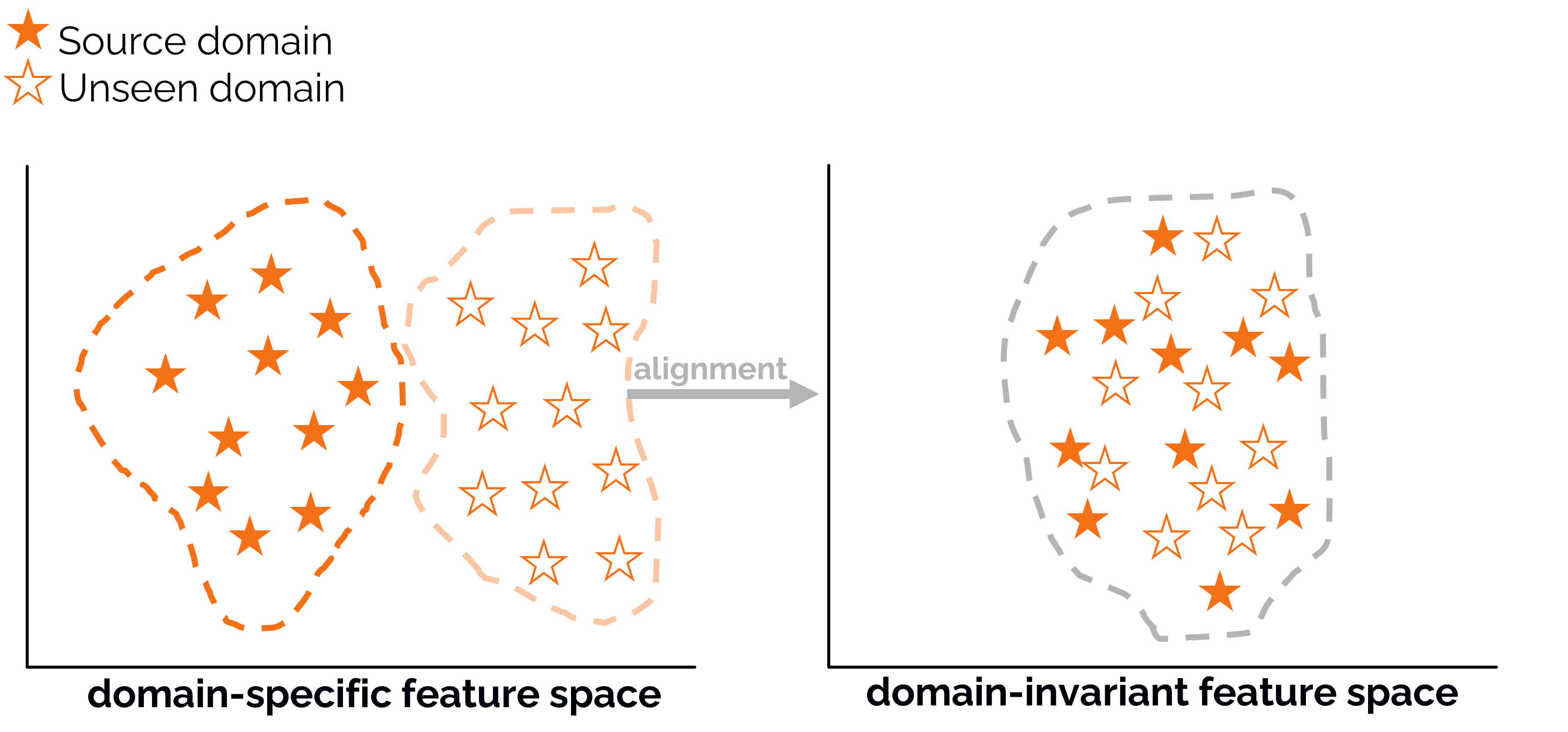} 
\caption{Alignment.} 
\label{fig:alignment}
\end{subfigure} 
\vspace{0.5em} 
\begin{subfigure}[t]{0.48\textwidth} 
\centering 
\includegraphics[width=\linewidth]{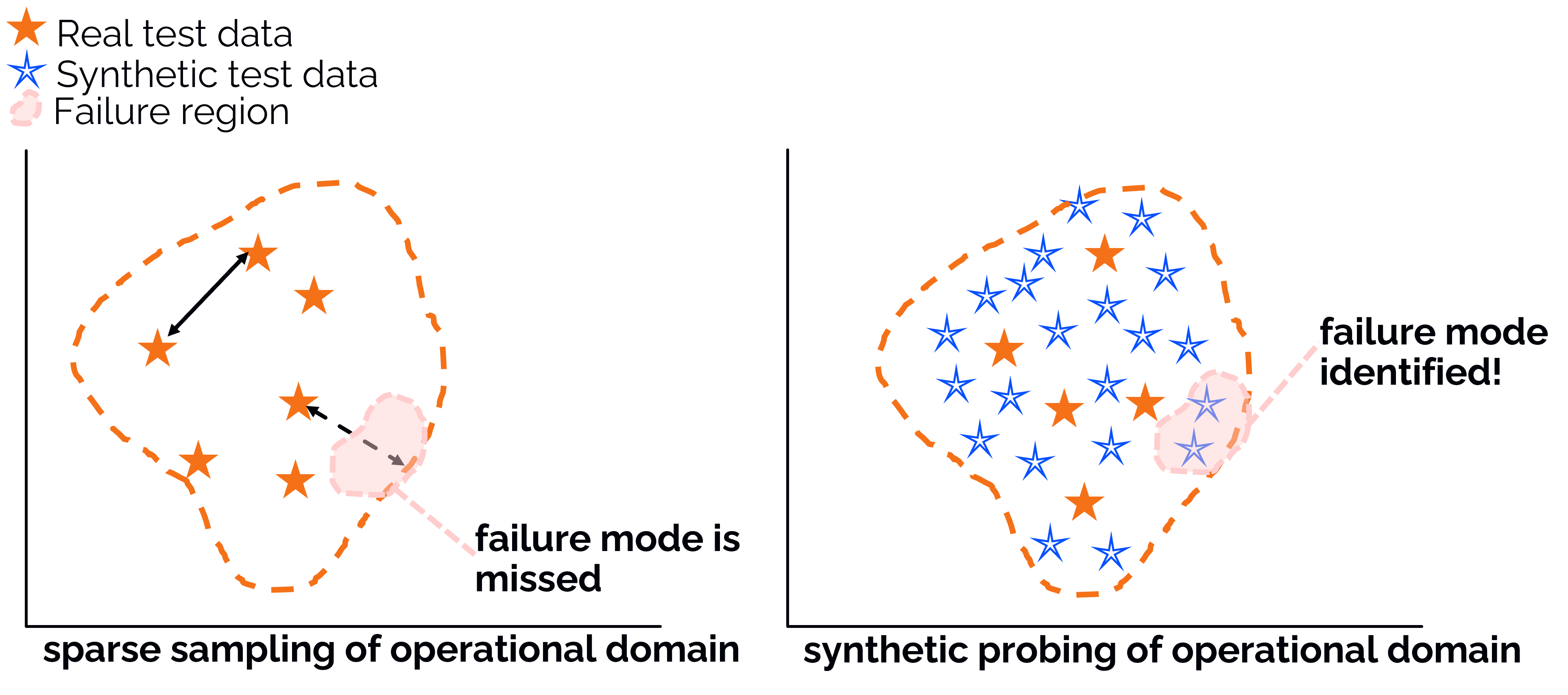} 
\caption{Synthetic data as a probe.} 
\label{fig:probe} 
\end{subfigure} 
\hfill 
\begin{subfigure}[t]{0.48\textwidth} 
\centering 
\includegraphics[width=0.8\linewidth]{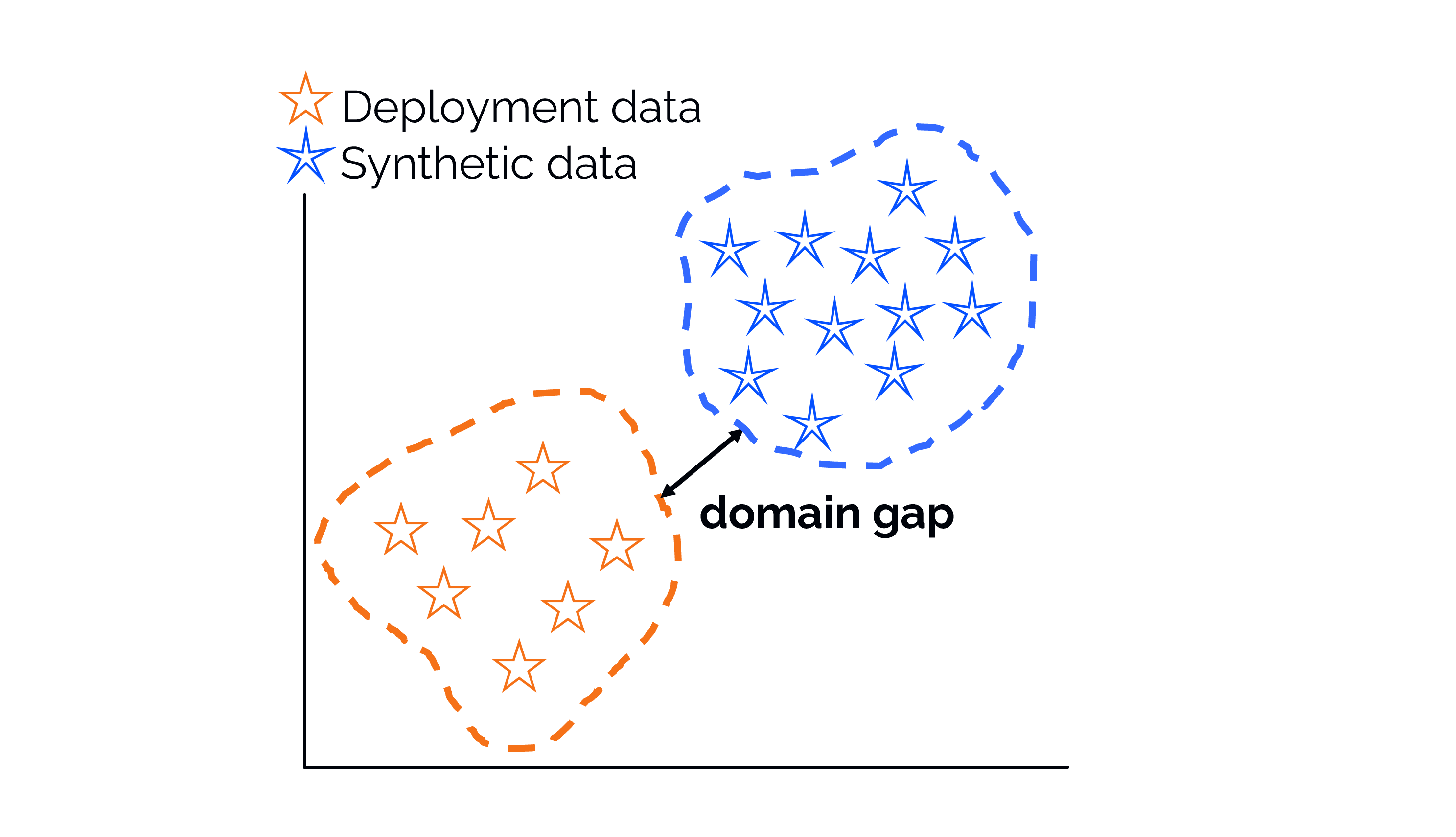} 
\caption{Synthetic-to-real gap.} 
\label{fig:gap} 
\end{subfigure} 
\caption{Overview of key domain generalization concepts and the role of synthetic data. (a) Diversification expands the support of the training distribution through augmentations and synthetic data generation, increasing coverage of unseen domains. (b) Alignment promotes domain-invariant representations by reducing the discrepancy between source and unseen domains in feature space. (c) Synthetic data can be used as a probe by enabling controlled and interpretable domain shifts for robustness evaluation and failure-mode analysis. (d) Synthetic-to-real transfer highlights the domain gap between synthetic training data and real deployment data, often resulting in degraded performance under deployment conditions.} 
\label{fig:illustration} 
\end{figure}

Many DG methods can be organized around two complementary principles: \textit{diversification}, which expands the support of the training distribution, and \textit{alignment}, which aims to reduce discrepancies between domains by learning domain-invariant representations. These concepts are illustrated in Figure \ref{fig:illustration}. 

\subsection{Diversification}

Diversification expands the support of the training distribution by increasing the diversity of the data available during training. This can be achieved by introducing variation in appearance, image statistics, or input conditions, thereby encouraging models to learn features that generalize beyond the training domain. Diversification approaches can broadly be grouped into feature-level and image-level methods, whereas additional strategies enabled by synthetic data are discussed in Section \ref{sec:syntheticenabler:diversification}.

\subsubsection{Feature-level}
The first approach to diversification is to manipulate the statistics of intermediate network features during training. The intuition is that domain shifts are often characterized by changes in image style, such as texture, color distribution, illumination, or sensor characteristics, while the semantic content of the image remains unchanged. By perturbing feature statistics, the model is exposed to a broader range of appearance variations and is encouraged to rely less on domain-specific cues. Unlike image-level approaches, diversity is introduced directly in the feature extractor, resulting in lower computational cost and less content distortion caused by image reconstruction artifacts. Representative methods include pAdaIN \cite{Nuriel2021}, DSU \cite{Li2022}, and MixStyle \cite{Zhou2024}. 

pAdaIN \cite{Nuriel2021} swaps activation statistics (channel-wise mean and standard deviation) between samples in a training batch. DSU \cite{Li2022} fits a Gaussian distribution model to the activation statistics in a batch and then generates new activations by random sampling. Fan et al. \cite{Fan2023} argue that since the above methods rely on inter-image variance in the training data set, they are incapable of significantly increasing the diversity in a typical single-source domain dataset. Instead, the authors propose changing the channel-wise per-image activity mean and standard deviation by a multiplicative factor sampled from a Gaussian distribution with mean one. They report significantly better results for OOD object detection when the method is applied to the first few layers of the detector feature extractor.

Another closely related approach concerns feature statistic mixing. MixStyle \cite{Zhou2024} synthesizes new domains by mixing feature statistics from two randomly selected instances from different training domains. Inspired by Adaptive Instance Normalization (AdaIN)\cite{Huang2017}, which was originally developed for image style transfer (see Section \ref{enabler:imagelevel}), MixStyle combines the channel-wise mean and variance of feature maps while preserving their semantic content. Exposure to such interpolated feature distributions encourages robustness to appearance variations across domains.

\subsubsection{Image-level}
\label{enabler:imagelevel}
 
Another approach to diversification is through image-level augmentations. Elementary single-image augmentations can be divided into photometric and geometric transformations. Photometric transformations include for example color jittering, blurring, sharpening, noise addition, brightness and contrast changes, and histogram equalization. They also include masking techniques such as CutOut \cite{DeVries2017}, Random Erasing \cite{Zhong2020}, and GridMask \cite{Chen2024}. Common geometric transformations include random rotation, flip, translation, scaling, and shear. Methods for optimizing sets of elementary image augmentations have also been proposed. For example, RandAugment \cite{Cubuk2020} automatically applies a small number of randomly selected augmentations from a predefined set. 

Diversification can also be achieved through learned image transformation. Style transfer methods generate new image appearances while preserving semantic content, thereby exposing the model to variations in texture, color, illumination, and other domain-specific characteristics. A widely used example is AdaIN \cite{Huang2017}, which transfers the style of one image to another by manipulating feature statistics. Similarly, DeepAugment \cite{hendrycks2021many} applies learned image-to-image transformation networks to generate realistic but strongly distorted training images, encouraging models to learn more robust representations under distribution shift.

It is also possible to combine multiple images. Originally developed for image classification, mixup \cite{Zhang2018mixup}  generates new samples by linearly superimposing two images and mixing their target labels using the same ratio. For object detection, the resulting image is typically assigned the union of the bounding-box and class annotations from both source images \cite{Zhang2019}. Thus, if image A contributes $70\%$ and image B $30\%$, objects from both images remain annotated in the same ratio for the mixed sample. Further improvements have been reported by interpolating the corresponding loss functions as well \cite{Vu2024}. Mosaic \cite{Bochkovskiy2020} augmentation combines several images by arranging them into a grid and extracting a random crop of the same size size as an individual image. The result combines parts of multiple images into one, reducing the spurious correlations between object and background, and increasing the statistical variation within a mini-batch \cite{vanLeeuwen2024SOD}. 

Beyond combining existing training samples, diversification can also be achieved by intentionally generating challenging perturbations. Adversarial training generates perturbed training images by applying gradient ascent on the loss function with respect to the image pixels. Zhang and Zhu \cite{Zhang2019a} show that training an image classifier on adversarial examples makes the network focus more on shape and less on texture, thereby improving robustness to distribution shifts. Several works have extended adversarial image augmentation to object detection to improve generalization \cite{Zhang2019b, Chen2021}.

Lastly, diversification can also be achieved by manipulating the frequency representation of images. Xu et al. \cite{Xu2023b} hypothesize that models relying more on phase information, which captures the structural content of an image, generalize better across domains than models that depend on amplitude information, which is more closely related to appearance and style. They compute the Fourier transforms of images from different domains and synthesize new training samples by randomly interpolating their amplitude spectra while preserving the corresponding phase information. The modified spectra are then transformed back to the image domain. Both the original and Fourier-augmented images are used during training to improve generalization. A related approach is Frequency-based Pseudo-Domain Generation (FPDG). \cite{zhangfrequency} FPDG generates pseudo-domains by decomposing images into structural phase and style-carrying amplitude spectra. The phase is kept intact to preserve bounding boxes, while the amplitude component is mixed. By exposing the detector to these synthesized pseudo-domains, it is encouraged to learn style-invariant features.

\subsection{Alignment}
While diversification increases variability in the training data, it does not explicitly encourage invariance to domain shifts. Although exposing a model to a broader range of appearances can improve robustness, unseen domains may still fall outside the support of the training distribution. As illustrated in Figure \ref{fig:alignment}, alignment instead aims to reduce the discrepancy between domains by learning representations that are insensitive to domain-specific characteristics.

In the context of object detection, alignment is particularly challenging because both localization and classification must remain robust under domain shift. Consequently, alignment may need to operate at multiple representation levels, including global image features and instance-level object representations. Existing approaches can broadly be categorized into adversarial, structure-aware, conditional, consistency-based alignment methods.

\subsubsection{Adversarial alignment}
Adversarial alignment aims to learn feature representations that are predictive for the task of interest while being invariant to domain-specific characteristics. A classical example is the Domain Adversarial Neural Network (DANN) \cite{Ganin2016}, which trains a feature extractor to support the target task and simultaneously to confuse a domain classifier that attempts to identify the source domain of each sample. The method can be extended to multi-source DG using a multi-class domain classifier \cite{Matsuura2020}. However, adversarial alignment is less straightforward for object detection than for classification. Zhu et al. \cite{Zhu2019} report only limited performance gains from applying DANN-based alignment to object detectors. They argue that global alignment is often dominated by background information because objects occupy only a small part of the image. Effective alignment for object detection should therefore focus more explicitly on object-related features \cite{Zhu2019,Lin2021,Seemakurthy2023,Danish2024}.

\subsubsection{Structure-aware alignment}
Structure-aware methods address this limitation by considering both global image representations and local object-level features. Lin et al. \cite{Lin2021} propose the Domain-Invariant Disentangled Network (DIDN), built on the Faster R-CNN detector, which explicitly separates domain-specific and domain-invariant information. Domain-specific encoders extract domain-exclusive features, while a shared encoder learns domain-invariant features used for detection. A decoder reconstructs each input image from both representations, while an adversarial domain classifier encourages the shared features to become domain-invariant. The same disentanglement is applied at the object-level by reconstructing region-of-interest feature maps.

\subsubsection{Conditional alignment}
Conditional alignment goes beyond matching the feature distributions across domains by also accounting for differences in task-specific outputs. Seemakurthy et al. \cite{Seemakurthy2023} argue that aligning the marginal feature distributions is insufficient, as the relationship between the extracted features and the predicted object classes and bounding boxes may remain domain-dependent. They therefore introduce a regularization term based on the Kullback-Leibler (KL) divergence between domain-specific conditional distributions, decomposed into separate classification and localization losses. This is combined with adversarial losses to enforce domain invariance and consistency regularization between global and instance-level domain classifiers. The authors extend this approach to one-stage detectors such as FCOS \cite{Tian2019}. In contrast to Faster R-CNN, this model processes outputs from multiple scales using a Feature Pyramid Network (FPN). Instance-level domain classifiers are applied across all anchor positions at all scales, while a global domain classifier is applied at the lowest resolution feature map. Focal loss is used to handle domain imbalance.

\subsubsection{Consistency alignment}
Building on conditional alignment, consistency-based methods encourage predictions or representations to remain stable across domains or different views of the same input.
Zhao et al. \cite{Zhao2020} use the KL divergence to penalize differences between class and bounding box predictions across domains. This explicitly aligns conditional distributions instead of only feature statistics. Danish et al. \cite{Danish2024} similarly enforce consistency between detections obtained from original and augmented versions of the same image, requiring both class probabilities and localization outputs to remain stable under perturbations. 

Contrastive learning provides an alternative to adversarial alignment by pulling together features extracted from different views of the same image while pushing apart features from different images \cite{Chen2023, xie2021detco}. In object detection, such consistency constraints are often applied at both the global image level and the instance or ROI level, highlighting the need for alignment across multiple representation levels.

\subsection{Vision-Language Model-based domain generalization}
\label{sec:VLM}
Recent DG methods make use of Vision-Language Models (VLMs). Rather than explicit diversification or alignment, these methods make use of semantic representations learned from large-scale image and text data. Such representations are less tied to domain specific visual characteristics, allowing them to improve robustness under distribution shifts. Due to the fact that VLM-based methods do not clearly fall under diversification or alignment, they are discussed separately.

The aligned representations of image and text provided by VLMs, e.g., CLIP \cite{radford2021}, enable semantic-level feature manipulation. In these approaches, the VLM image encoder essentially serves as the detector backbone network, with additional perturbation, normalization, or adaptation modules operating on its features. Vidit et al. \cite{vidit2023} and Fahes et al. \cite{fahes2023} estimate semantic augmentation parameters from text prompts to transform source image features into domain-specific representations, including different weather conditions and time of day. Zhang et al. \cite{zhang2025style} extend this idea by generating composite style augmentations from multi-step (chain-of-thought) text prompts. These approaches use VLMs to introduce semantically meaningful style and domain variation into the training data. In contrast, Zhang et al. \cite{Zhang2026CVPR}, use VLM-derived domain representations to explicitly encourage image embeddings to become orthogonal to domain-specific information, thereby promoting domain-invariant representations. A limitation of these approaches for non-RGB modalities, e.g., thermal imagery, is that the VLM image encoder and its aligned image–text representation are primarily learned from RGB imagery. Consequently, the semantic alignment and domain representations may be poorly matched to characteristics specific to other modalities.

\section{Synthetic data as enabler for domain generalization}
\label{sec:syntheticenabler}
The previous section reviewed DG approaches that improve robustness to unseen domains through diversification and alignment. In this section, we revisit these two principles from the perspective of synthetic data. When used alongside real-world training data, synthetic data can support diversification by expanding the variability of the training distribution and support alignment by reducing discrepancies between domains through the generation or transformation of data.

\subsection{Diversification}
\label{sec:syntheticenabler:diversification}

Synthetic data provides a natural mechanism for image-level diversification by enabling controlled generation of additional training samples under varying conditions. For example, Heslinga 
et al.\cite{Heslinga2024combining} show that supplementing a small set of real images with a large set of simulated military vehicle images improves object detection performance. The same study demonstrates that increasing variation within the simulated dataset leads to additional gains in detector performance. 

More recently, the role of generative models in synthetic data generation was reviewed by Fokkinga et al. \cite{fokkinga2025generative}, who distinguish between full-image generation, inpainting, and image-to-image translation approaches. Full-image generation synthesizes complete scenes from prompts or conditioning inputs and has been used to create training data for domains where real imagery is scarce. For example, class-specific diffusion models were shown to improve military vehicle detection in low-data regimes by generating additional training images from only a handful of real examples \cite{Fokkinga2026-or}. Similarly, Malizia et al. address data scarcity in natural environments by using a VLM to prompt Blender scene generation from reference photos, synthesizing ground-truth-labeled training scenes \cite{malizia2026diffuseobjectlabelcurating}. 

Inpainting methods insert new objects into existing scenes while maintaining consistency with the surrounding environment. Snel et al. demonstrated that diffusion-based object inpainting can improve vehicle detection performance in UAV imagery by augmenting minority classes with synthetically inserted vehicles \cite{snel2025data}. Image editing methods modify existing images through textual or structural guidance, enabling controlled variation of object appearance, pose, or environmental conditions while preserving scene layout \cite{Stein2026}. Finally, image-to-image translation methods, such as GAN- or diffusion-based style transfer approaches, transform images between domains (e.g., synthetic-to-real, day-to-night, RGB-to-infrared)\cite{Eker2026} and are widely studied to reduce visual domain gaps in object detection pipelines. 

A distinguishing characteristic of foundational generative AI models is the ability to combine functionalities within a single framework. Modern diffusion models support text conditioning, structural control through methods such as ControlNet \cite{zhang2023adding}, object-level editing, and inpainting, allowing synthetic datasets to be tailored to specific detection challenges. These capabilities make generative models an increasingly attractive alternative to traditional simulation pipelines for improving object detector robustness and DG.

Synthetic imagery may, however, also exhibit characteristic statistical deficiencies that limit its usefulness for diversification. For example, PASTA \cite{Chattopadhyay2023} observes that synthetic images tend to have smaller variations in the high-frequency components of their amplitude spectrum than real images, likely due to their more limited texture variation. PASTA addresses this mismatch by applying random Gaussian perturbations to the amplitude spectrum in the Fourier domain while preserving the phase information, thereby increasing the frequency-level diversity of synthetic training data. This significantly improves synthetic-to-real generalization for object detection, especially when combined with RandAugment. \cite{Cubuk2020}

\subsection{Alignment}
In addition to supporting diversification, synthetic data can be used to facilitate alignment between domains. Recent work has explored the use of generative models to transform samples into a shared intermediate domain, thereby reducing domain discrepancies. Unlike conventional alignment approaches, which operate on features or predictions, these methods perform alignment directly in the image space.

A notable example is Diffusion-Driven Adaptation (DDA) proposed by Gao et al. \cite{GaoTTA}. In contrast to conventional Test-Time Adaptation (TTA) methods, which adapt model parameters using target-domain data, DDA keeps the source model fixed and instead adapts the input. A diffusion model trained on source-domain data is used to project target-domain data towards the source domain, reducing the domain gap prior to inference.

While DDA demonstrates that generative models can effectively align target data towards the source domain, subsequent work has argued that the resulting images may still form a distinct synthetic domain from the perspective of the downstream network. Guo et al. \cite{guo2025everything} therefore propose Synthetic-Domain Alignment (SDA), which aligns both the source model and target data to a shared synthetic domain. Their approach first fine-tunes the source model using diverse synthetic images and subsequently projects target-domain samples into this same synthetic space at test time. This allows SDA to achieve stronger domain consistency and improved robustness under distribution shift.

\section{Synthetic data as probe for domain generalization}
\label{sec:syntheticprobe}
% \textbf{What synthetic data reveals about DG}

% \subsection{Probing for domain generalization}
The previous sections considered synthetic data as an \emph{enabler} of DG. However, synthetic data can also serve a second, complementary role: as a \emph{probe} for understanding DG itself. As illustrated in Figure \ref{fig:probe}, synthetic data enables controlled and interpretable domain shifts through systematic variation of factors such as scene composition, illumination, weather conditions, and sensor characteristics. This allows for carefully designed experiments that isolate individual sources of variation and analyze their effect on model performance. Despite its potential, the use of synthetic data for validation, verification, and probing of DG remains considerably less explored than its use for training.

Van Breugel et al. \cite{vanbreugel2023} address this gap by proposing the use of synthetic test data to improve model evaluation under data scarcity and distribution shift. Their framework leverages generative models to create test samples in underrepresented regions of the data distribution and to simulate plausible deployment shifts, enabling more reliable estimation of model performance than what can be obtained from limited real-world test sets alone. Their results demonstrate that synthetic test data can improve performance estimation for minority subgroups and shifted deployment scenarios. Closely related work\cite{fokkinga2024validation} investigates the use of simulation as a surrogate for real-world test data in the evaluation of object detectors. Results suggest that simulated  environments can reproduce many performance trends observed in real-world testing, supporting their use as a tool for evaluation. However, the validity of conclusions drawn from synthetic evaluations ultimately depends on the representativeness and fidelity of the simulation itself.

Synthetic data enables the creation of interpretable test conditions through controlled manipulation of factors such as illumination, weather, blur, noise, and background appearance. By varying a single factor while keeping all others fixed, nuisance variables can be isolated and changes in detector performance can be directly attributed to specific sources of domain shifts. Such perturbation studies have been used to systematically assess object detector robustness to environmental changes and adverse operating conditions \cite{gupta2024robust,premakumara2023,Nijskens2023PredictingDNN}. 

Beyond robustness evaluation, synthetic data also enables the generation of rare or safety-critical scenarios that are insufficiently represented in real-world test sets. Such scenarios may include extreme environmental conditions, scarce assets, unusual object configurations, severe occlusions, sensor artifacts, or other edge cases that occur too infrequently for reliable empirical evaluation. By exploring these regions of the operational domain, synthetic test data can reveal failure modes and characterize the conditions under which performance degrades. While this capability has received limited attention in the object detection literature, similar ideas have been explored in other domains. For example, diffusion-based image editing has been used to simulate controlled distribution shifts and diagnose failure modes of biomedical vision models \cite{perez-garcia_bond-taylor_radedit}. 

\section{The synthetic-to-real gap in object detection}
\label{sec:syntheticgap}
% Synthetic data for object detectors 
The preceding sections reviewed how synthetic data can be used both to improve DG and to probe model behavior under controlled shifts. Another important use case of synthetic data is the training of object detectors, either exclusively on synthetic imagery or in combination with real-world data. Several examples of such approaches were discussed previously in the context of image-level diversification (Section \ref{enabler:imagelevel}). Synthetic data has become a widely used alternative to real-world data for training object detectors, with applications including radar object detection, UAV remote sensing, and drone detection \cite{7738056,bialer2024radsimrealbridginggapsynthetic,rs18020361,electronics12102197}. However, despite the potential of synthetic data to be used in scarce data domains and to increase training diversity, models trained on it often experience a significant performance degradation when deployed on real-world imagery.

% The gap 
This degradation is commonly referred to as the \emph{synthetic-to-real gap}, \emph{simulation-to-real gap}, or \emph{reality gap} (Figures \ref{fig:overview_dg} and \ref{fig:gap}). The gap arises from differences between synthetic and real-world data distributions, including object appearance, environmental conditions, sensor characteristics, background complexity, and scene composition. As a result, transfer performance can be strongly affected by factors such as lighting, occlusion, texture, object geometry, and the presence of previously unseen distractors \cite{10.1007/978-3-030-87156-7_9,Sudhakar_2023_ICCV,electronics12102197,rs18020361}. From a DG perspective, the synthetic-to-real gap can be viewed as a specific instance of domain shift. Synthetic-to-real transfer is concerned with training models on synthetic data in such a way that the model is also effective on real-world data. The remainder of this section discusses the factors that influence synthetic-to-real transfer in object detection.

\subsection{Data-centric strategies for closing the gap}
% how can we reduce it -> domain randomization 
One of the main approaches to reduce the synthetic-to-real gap is through domain randomization. Domain randomization can be viewed as a particularly aggressive form of diversification (Section \ref{sec:syntheticenabler:diversification}), in which synthetic data are deliberately varied along numerous factors such as lighting, texture, background appearance, and object pose. As an example, Yu et al. \cite{Yu2021} show that larger and more diverse pretraining datasets improve both in-domain and OOD performance for image classification models, with similar improvements observed when model size increases. Domain randomization does not necessarily focus on accurately reproducing a specific deployment environment, but the goal is to expose the model to such a wide range of variability that the real-world domain falls within the support of the training distribution. Especially if there is extensive variation in the aforementioned factors, studies demonstrate that domain randomization approaches are effective \cite{8575297,zhu2025domainrandomizationobjectdetection}. Other methods that were discussed in Section \ref{sec:syntheticenabler} such as PASTA and MixStyle, can also be effective in this scenario.

% diversity > realism
Interestingly, multiple studies indicate that increasing variation, including through feature-based diversification methods, is often more effective for improving synthetic-to-real transfer than increasing visual realism, even though the latter is frequently the focus of generative AI approaches \cite{arayamartinez2025syntheticindustrialobjectdetection,8575297}. From a DG perspective, this suggests that the ability of synthetic data to represent a broad range of deployment conditions may be more important than accurately reproducing a single real-world domain. Furthermore, some aspects of real-world variability may be inherently unpredictable and therefore impossible to model explicitly \cite{aljalbout2025realitygaproboticschallenges}. In other words, there will always be unknown or unanticipated scenarios or variations, which could indicate that fully closing the gap may not be possible.

Finally, studies have observed that adding even a small amount of real-world data to a synthetic dataset can give substantial improvements in detector performance. Heslinga et al. \cite{Heslinga2024combining} demonstrated that combining synthetic images with a limited number of real samples significantly improved fine-grained military vehicle detection, while Moon et al. \cite{moon2025synaug} reported similar benefits when combining synthetic and real data to address data imbalance. A possible explanation as to why this works is that real data corrects shortcut learning, which could be caused by discriminative features that are actually artifacts in the synthetic data. Ljungqvist et al. \cite{Ljungqvist2023} provide empirical evidence that detectors trained on real or synthetic data differ mainly in later, high-level representation layers. A small amount of real data could therefore alter how local cues are combined in these layers without requiring the network to relearn underlying low-level features, which would require much larger amounts of data. This observation further supports the view that synthetic-to-real transfer is not only a problem of data realism or diversity, but also one of representation learning.

\subsection{Model-centric strategies for closing the gap}
While data-centric strategies aim to reduce the discrepancy between synthetic and real imagery, they do not fully address the challenges of synthetic-to-real transfer. Generalization performance is also influenced by model scale, pretraining data, and detector architecture. Zhang et al. \cite{Zhang2022} show that vision transformers exhibit improved robustness to distribution shifts compared to convolutional networks. This trend has been observed in several other studies \cite{Bai2021,Paul2022,Zhou2022}.

% Model architecture - object detection gap 
Similar observations can be found in object detection, where generalization performance varies substantially across detector architectures. Among studies addressing DG in object detection, two-stage detectors, mainly Faster R-CNN \cite{Ren2015}, dominate. This may be explained by the similarity between the second-stage ROI classification module and conventional image classification architectures. However, Faster R-CNN is generally not competitive for real-time video processing when compared to modern one-stage detectors, particularly those from the YOLO family \cite{Redmon2016}. More recent transformer-based detection algorithms, such as DETR \cite{Carion2020} and ViTDet \cite{Li2022a} have achieved state-of-the-art performance on in-domain benchmarks. ViTDet models that are pre-trained on massive datasets using self-supervised learning show strong OOD performance \cite{He2022}. However, when training on smaller datasets, OOD performance degrades substantially as reported by Mao et al. \cite{Mao2023}. These authors also find that DETR has comparatively poor OOD performance, possibly due to deficiencies in the matching of predictions to ground truth boxes. These findings highlight that strong in-domain performance does not necessarily translate to improved robustness under domain shift. Both dataset characteristics and detector architecture influence synthetic-to-real transfer, suggesting that closing the gap requires a combination of data-centric and model-centric approaches.

There are several types of model-centric approaches that can be applied to reduce the synthetic-to-real gap, although there is little research specifically investigating this setting. One approach is meta-learning, where the goal is to learn a model in such a way that the model can quickly adapt to new domains. Meta-learning techniques that apply more broadly to DG include MLDG (Meta Learning for DG \cite{Li2018a} and Meta Regularization (MetaReg) \cite{Balaji2018}. In MLDG, multiple train and test sets exist from different domains and during each iteration these are mixed, mimicking a scene where the test domain is different from the training domain. Whereas this MLDG approach focuses on learning better model parameters, MetaReg focuses on learning a better training objective. Another approach, also discussed in Section \ref{sec:VLM}, is CLIP-the-Gap \cite{vidit2023}, which is more a semantic representation learning approach. By using the semantic priors from a pretrained CLIP model, a detector model can be encouraged to learn domain-invariant representations of the objects. This could improve generalization of an object detection model to real-world scenarios, even though it was partly trained on synthetic images.

\section{Conclusion}
\label{sec:conclusion}
% DG is still a fundamental challenge for object detection, many methods from classification and do not transfer to detection.
DG remains a significant challenge for object detection. Although foundational models have improved robustness under distribution shift, substantial performance degradation persists, particularly in data-scarce domains. Unlike image classification, object detection requires robustness for both localization and classification, and domain shifts can affect these components differently. As a result, approaches developed for classification often fail to fully address the challenges of object detection.

% diversification + limitations AND alignment + limitations
This review highlights several limitations of current DG approaches. Diversification expands the support of the training distribution through augmentations, synthetic data generation, and domain randomization. However, improved coverage of the training distribution does not guarantee coverage of deployment conditions, as it remains difficult to anticipate which variations will be encountered in in real-world data. Therefore, there is no assurance that such an approach will lead to robust generalization. Similarly, the approaches based on alignment show important limitations when applied to object detection. The explicit feature normalization, adversarial learning, or conditional based techniques, aim to make features from the network domain invariant. However, this is a more complex process in detection tasks than in classification tasks, as invariance should be imposed on both global and local levels to address domain shifts in overall style and object local structure, respectively.  Consequently, approaches developed for classification often fall short when applied to object detection.

% gap 
The synthetic-to-real gap provides a particularly relevant case study for DG. Although synthetic data offers a scalable and time-efficient way to increase both the quantity and diversity of training data, substantial performance degradation can still occur when models are deployed on real-world imagery. This suggests that closing the gap requires more than increasing realism or variation alone; it also requires understanding how domain shifts affect the representations learned throughout the detection pipeline. An important open question is why even small amounts of real-world training data often lead to substantial improvements in synthetic-to-real transfer. Possible explanations include sensor or scene modeling inaccuracies, spurious correlations unique to synthetic imagery, or mismatches in the distribution of target signatures. This understanding of how representation are affected could be further investigated through the use of synthetic data as a probe, where it can enable controlled experiments that isolate these factors and quantify their impact on detector performance. This positions synthetic data both as a practical DG challenge in synthetic-to-real transfer, and as a tool for systematically studying generalization and failure modes.

% conclusion
Together, these observations suggest that future DG research should move beyond directly transferring methods from image classification and instead develop representation-aware approaches specifically for object detection. Such methods should explicitly account for the more complex nature of a detection model, where both global and local features are of importance. Recent progress in VLMs further supports this direction, as their learned semantic representations appear to offer improved robustness under domain shift. When such semantic priors with detector-specific localization mechanisms and synthetic-data-driven training strategies are integrated, this could provide a promising route towards more generalizable object detectors. In addition, it is necessary to further explore the impact of using synthetic data as tool for probing domain shifts and failure modes, and more generally for validating detection models. The combination of these approaches may improve not only the robustness of object detectors, but also our understanding of why they succeed or fail under distribution shift.

% References
\bibliography{report} % bibliography data in report.bib
\bibliographystyle{spiebib} % makes bibtex use spiebib.bst

\end{document}